# A new method color MS-BSIF Features learning for the robust kinship verification

Rachid Aliradi
*Dept. of DRDHN*
CERIST
Algiers, Algeria
roliradi@gmail.com

Abdealmalik Ouamane
*Dept.of Electrical Engineering*
University Mohamed Khider
Algiers, Algeria
ouamaneabdealmalik@yahoo.fr

Abdeslam Amrane
*Dept. of DSISM*
CERIST
*Algiers, Algeria*
amraneabdeslam@gmail.com

*Abstract*—the paper presents a new method color MS-BSIF learning and MS-LBP for the kinship verification is the machine's ability to identify the genetic and blood the relationship and its degree between the facial images of humans. Facial verification of kinship refers to the task of training a machine to recognize the blood relationship between a pair of faces parent and non-parent (verification) based on features extracted from facial images, and determining the exact type or degree of this genetic relationship. We use the LBP and color BSIF learning features for the comparison and the TXQDA method for dimensionality reduction and data classification. We let's test the kinship facial verification application is namely the kinface Cornell database. This system improves the robustness of learning while controlling efficiency. The experimental results obtained and compared to other methods have proven the reliability of our framework and surpass the performance of other state-of-the-art techniques.

## I. INTRODUCTION

These days, facial kinship verification (KV) by computer is very rich research subject and especially for the analysis of facial images which is considered as an attractive and interesting field of research study for the field of computer vision and image processing. But remains an essential and for visual recognition open problem. This consideration is due to the availability of many socially relevant information about the human face, such as gender, age, and emotional state and social media.

Checking if two people come from the same family or not. The face is the important that people use to recognize themselves. Humans can quickly to identify each other by their the facial features and, the face remains as a dynamic recognition approach to the man.

The KV is the machine's ability to identify the genetic and blood relationship and compare its degree between the facial images of humans proposed by Shang et al. [1]. Face texture and discriminating kinds and traits features are a free natural password and a means through which people recognize each other, it is used in the early stages of life by an infant to recognize his surrounding people Proposed by Aliradi et al. [2]. The KV refers to the task of training a machine to recognize the blood relationship between a pair of faces parent and non-parent (verification) based on features extracted from facial images, and determining the exact type or degree of this genetic relationship. Automatic kinship verification is an interesting area of inspecte and has an impact significant in many real-world applications, e.g. member search missing family members, historical research and genealogy, image annotation, creation of family trees, search for missing children and forensics, are targeted by kinship verification.

Further, the face attributes of humans of the same family may indicate a large dissimilarity whereas pair faces of persons with no kinship may look similar. All these challenges grow the difficulties of the KV problem. There are real problems when the discriminative features are used in traditional kernel verification systems, such as concentration on the local information zones, containing enough noise and redundant information in zones overlapping in certain blocks, manual adjustment of parameters and dimensions high vectors. To solve the above problems, Oualid et al. Proposed [8]. A new method of robust face verification with combining with a large scales local features based on Discriminative-Information based on Exponential Discriminant Analysis (TXQDA).

Motivated by his research, communities of vision computer and automatic learning manifest an increasing interest in incubation and the promotion of calculation methods to verify the relationship between humans to from face images.

Kin directly linked to one another, such as mother-child, father-child, and sister-brother, is called primary kinship. Secondary kinship means the primary kin of the first-degree kin, such as uncles, sister's husband, and brother's wife. Tertiary kinship refers to the primary kin's secondary kin like the brother of sister's husband as in Almuashi et al. [3]. Zhang et al. [5] developed local operators. On first the decomposed each image into different non-overlapping blocks. Next, they tried to derive a matrix to project the bloks in order an optimal subspace to maximize the different margins of different individuals. Each column was then designed to an image filter to treat facial images and the filter responses were binarized using a fixed threshold. In fact, we can say that facial kinship verification systems can be affected by several influences such as: lighting, occlusions, emotions, facial expression, makeup, cosmetic surger. Zhang et al. [6] introduced a novel Nearest Neighbor Classifier (NNC) distance measurement to resolve easy SSFR problems. The suggested technique, entitled Dissimilarity-based Nearest Neighbor Classifier (DNNC), divides all images into equal non-overlapping blocks and produces an organized image block-set. The dissimilarities among the given query image block-set and the training image block-sets are calculated and considered by the NNC distance metric.

Additionally, kinship verification adds another dimension layer of difficulty which is from solution require is far being to resolve easy. Our subject therefore concerns the verification of kinship, and we propose an extension of the learning of the descriptor BSIF (Binarized Statistical Image Features) and we combine and apply the local descriptors LBP (Local Binary Pattern) to extract features in order to study the influence of these on the performance of our kinship verification system. We propose in this paper a new method Color MS-BSIFeatures Learning and MS-LBP for robust kinship verification. We use the LBP and color BSIF descriptors for the comparison and the TXQDA method for dimensionality reduction and data classification. We let's test the facial kinship verification application on the Cornell kinship database is proposed by Duan, Q et al. [4].

Our contributions in this paper are therefore as the following:
1) We proposed a new approach, which is called color BSIF at feature level or scores (our work is the first that uses this method in the field of facial verification of kinship, to use the
advantage of color BSIF, and LBP methods. This approach uses multiple descriptors as well as based on features extracted from facial images, and to compute the degree similarity of that relation (verification) for pairs (match)- verified/ or (No-match) verified of faces kinship.
2) We observe the performance change with variation of rays (R) and the filter size (L), for the LBP, BSIF and the color BSIF learning descriptors.
3) There is a great improvement in the accuracy by the apprentice kinship descriptor application BSIF on different local descriptors.
4) We used cosine similarity measure after discriminant analysis phase to minimize the classification error.
5) Finally, the results proved that both of the two initializations are efficient. In addition, our method realizes the state-of-the-art performance on the problem of face verification and kinship (see Section [IV]). The remainder of this paper is organized as follows. (See Section [II]) presents the proposed facial verification of kinship approach. (See Section [III]) shows the experimental results. Finally, (see Section [V]) presents the conclusion of this paper.

II. THE PROPOSED FRAMEWORK FOR KINSHIP VERIFICATION

In this section, we present a framework for a facial kinship verification. Fig 1. Shows the principal building blocks of the proposed system. The proposed system enclose diverses components including face image pairs preprocessing, local facial feature extraction by learning color BSIF and LBP, projection per feature dimensionality reduction (TXQDA), cosine similarity measurement, and decision. In the following, we describe the details of each component

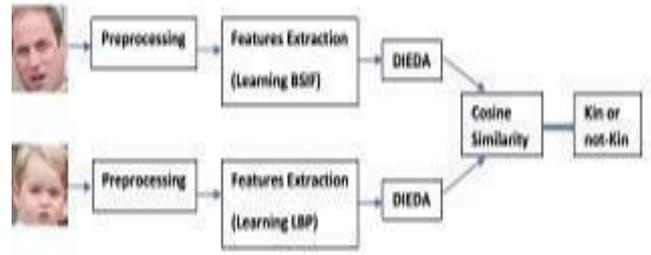

Fig.1 Overview of the Proposed Framework for Kinship Verification

A. *Preprocessing*

B. *The preprocessing of the image consists in preserving the maximum of the variations intrinsic of the face, and to remove the other information like the noise, background, the hair, shirt collars, ears...etc. in order to improve the performance of the facial verification of kinship. A rectangular window of size (64 × 64), centered on the most stable features related to the eyes, eyebrows, nose and mouth, was used.*

C. *Feature Extraction by BSIF Descriptor Learning.*

For feature extraction in our face verification application of the kinship we use our BSIF Learning Descriptor for facial verification of parentage. Our work is the first that uses this method in the field of facial verification of kinship. We are using also the LBP and BSIF descriptor for the comparison.

D. *BSIF descriptor learning.*

The BSIF descriptor was recently proposed by Kannala et al. [8] for the extraction of characteristics of 2D textures and face images. The set of filters used by the original BSIF descriptor [46] are constructed based on 13 natural images. In our work, we propose the use of the face color training images cut out for the construction of these filters. The color images cut in the learning phase are first made of zero mean (the mean value of each image is subtracted) and are divided by their deviation- type. We randomly select 50000 blocks for the construction of the filters (Red, Green, and Blue). The filters obtained with the size (L= 9) and the number of bits (n= 8) with the use of cut-out color images from the Cornell KinFace database. To our knowledge, it is the first work that uses the cut-out color images for the construction filters used by the BSIF descriptor. The corresponding color BSIF Learning code is presented in Figure 2.

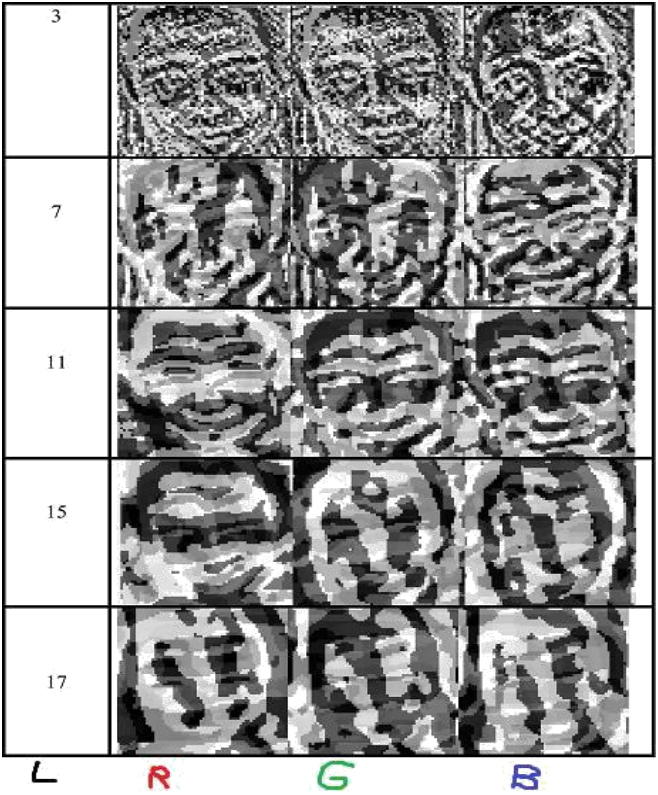

*Fig 2. BSIF descriptor learning of a color image for multiple values of (l).*

### E. Reduction and classification.

*After the feature extraction step and the concatenation of the 16 histograms of each block in a vector that forms the features of the faces. We use the method TXQDA for the reduction and classification of these characteristic vectors.*

*Algorithm 1 explain below shows the steps for implementing the method TXQDA.*

*Input: The X Database.*

*Output: the transformation matrix WTXQDA*

1. Compute the matrices $Sb^{sild}$ and $Sw^{sild}$ (as defined in [7]).

2. Compute the matrices $expb^{sild}$ and $expw^{sild}$.

3. Compute the eigenvectors and the corresponding Eigenvalues $(expw^{sild})^T expb^{sild}$ )

4. Compute the TXQDA transformation matrix which the first m eigenvectors ($W^{TXQDA}$).

### F. Cosine Similarity Metrics.

*All features extracted from each pair of two images should be integrated into a single value.*
*Our experiments examine in the literature different metric measures (Euclidean, Hamming, Manhattan, Cosine etc.) demonstrated that cosine similarity metric [8] provides the best performance.*

*We choose the cosine similarity measure after the reduction and classification of the characteristic vectors. The method* calculates the cosine score between the two vectors features ($x_1$ and $x_2$) are given by:

$$Scos(x_1, x_2) = \frac{((W^{TXQDA})^T x_1 (W^{TXQDA})^T x_2)}{\|(W^{TXQDA})^T x_1\| \|(W^{TXQDA})^T x_2\|}$$

Certainly, ||.||. The Euclidean norm. A high value of the produced score means a high probability that $x_1, x_2$ are same family.
With: $W^{TXQDA}$ the projection matrix of the TXQDA method

### III. EXPERIMENTAL ANALYSIS

For experimental evaluation, we considered the largest Cornell KinFace [8] includes 150 pairs of images of people's faces with kinship relationship. The database was collected from the Internet considering four family relationships. The distribution of family relationships in the dataset is 40% father son (P-Fs), 22% father-daughter (P-F), 13% mother-son (M-Fs) and 25% mother-daughter (M-F). Each pair is made up of a parent face image and a child face image. In our experiments, we evaluate the proposed approach on the same protocol as defined by Yan et al [13], where only 143 pairs are used and classification is performed with 5 cross-validations for kinship verification. Negative pairs are generated randomly by associating each parent image with an image of another child.

### A. Experimental Setup

The number of the negative and positive pairs used in the experiments is the same for each relation on the four subsets. We evaluate the mean accuracy over the five folds. The negative pairs and folds are predefined for the all four relations. For the extraction local features, we extracted color LBP and color BSIF-learning. For color LBP, the rays size is R = {1, 2, 3} for each face sample. For color BSIF, we use eight filters with different sizes L = {3, 7, 11, 15, 17} for each face sample. We calculate the computational complexity analysis of our system. The number is determined empirically and set to 10.

We used 5-fold cross-validation technique on the kinface Cornell dataset and we evaluate the mean accuracy over the five folds. We considered each pair of face images with kin relation as a positive sample and those without kin relation as negative samples. For kinship verification, the number of positive samples is often much smaller than that of the negative samples. To substantially exploit the relative characteristics encoded in kin data, TXQDA works by generating negative samples by using the LR-Fusion face images of the parent (or child) face images from each subset. Since our proposed TXQDA method works under the image-restricted setting, we compare and match our new method with other state-of-the-art algorithms methods conferring to the image-restricted protocol on the two kin dataset.

### B. Results and analysis

We run the experiments on the four relations of the kinface Cornell databaseThe results of these experiments are reported in Table I. As can be noticed from the latter, the performances change with the variation of rays (R) and filter size (L) of the LBP, BSIF and Color BSIF Learning descriptors and there is great improvement in accuracy by applying color BSIF Learning descriptor on different local descriptors.

The best result obtained by applying the Color BSIF Learning descriptor (L=3) with precision=80.53%. The performance of color MS-BSIF learning is much better than that of others methods in all cases our proposed method is compared against

C. *Conclusion and Perspective.*

In this paper, we proposed an approach that handle facial kinship verification problem. We presented a verification system that is based on integrating two different types of extracted features with different multi-scales. The system learns to decide whether two persons are with a kinship or not.

We are interested in the problem of kinship verification. Our work consisted in the development of a robust algorithm
intended to recognize the relationship of the kinship between two individuals by their faces. The feature extraction method Learning Color BSIF descriptor from color facial images. The BSIF method calculates a binary code string for the pixels of a given image. The code value of a pixel is considered as a local descriptor of the image intensity pattern, in the neighborhood of the pixel. In addition, the histogram of pixel code values allows to characterize texture properties in image regions and color information also remains an effective factor in our kinship verification system. The proposed BSIF Descriptor Training method uses color training images of the cutout face for the construction of these BSIF filters. Experimental results on kinface Cornell dataset show that our Color MS-BSIF and LBP learning method outperformed other state-of-the-art methods in Table1. As future work, we plan to investigate the complementarity of more features description for facial kinship verification representation with the proposed method. Apply our system to another database larger than the Cornell Kinship database and to applique machine and deep learning.
Merge multiple filter sizes (L) of our proposed descriptor Color BSIF Learning at feature level or scores.

| Method | Mean |
|---|---|
| **ResNet+CF [12]** | 65.51 |
| **SphereFace [11]** | 65.60 |
| **ResNet+SDMLoss [10]** | 65.58 |
| **Deep-Tensor+ELM [9]** | 68.61 |
| **Color MS-BSIF Learning** | 80.53 |

Table I: Averaged verification accuracy scores (%) on Kinface Cornell database for All relations.